# LLMs are Not Just Next Token Predictors

**Alex Grzankowski**
Reader in Philosophy
Birkbeck, University of London
School of Advanced Study, Institute of Philosophy, University of London
a.grzankowski@bbk.ac.uk
*corresponding author

**Stephen M. Downes**
Professor of Philosophy
University of Utah
s.downes@utah.edu

**Partick Forber**
Professor of Philosophy and Department Chair
Tufts University
patrick.forber@tufts.edu

**Abstract**
LLMs are statistical models of language learning through stochastic gradient descent with a next token prediction objective. Prompting a popular view among AI modelers: LLMs are just next token predictors. While LLMs are engineered using next token prediction, and trained based on their success at this task, our view is that a reduction to just next token predictor sells LLMs short. Moreover, there are important explanations of LLM behavior and capabilities that are lost when we engage in this kind of reduction. In order to draw this out, we will make an analogy with a once prominent research program in biology explaining evolution and development from the genes eye view.

LLMs are statistical models of language learning through stochastic gradient descent with a next token prediction objective. So, LLMs are 'just next token predictors', a popular view among AI modelers, explicitly laid out by Shanahan (2024): "A great many tasks that demand intelligence in humans can be reduced to next-token prediction with a sufficiently performant model" (2024, 68), and "surely what they are doing is more than 'just' next-token prediction? Well, it is an engineering fact that this is what an LLM does. The noteworthy thing is that next-token prediction is sufficient for solving previously unseen reasoning problems" (2024, 77). Yet LLMs won't satisfy users if they merely produce just any next token. The desired next token must contribute to answering questions, forming paragraphs, giving advice, making jokes, and so on. From the perspective of the user, LLMs are more than just next token predictors – they are joke tellers, sources of advice, and so on. Which perspective is the right one?

While LLMs are engineered using next token prediction, and trained based on their success at this task, our view is that a reduction to "just next token predictor" sells LLMs short. Moreover, there are important explanations of LLM behavior and capabilities that are lost when we engage in this kind of reduction. In order to draw this out, we will make an analogy with a once prominent research program in biology explaining evolution and development from the "gene's eye view".

Our position can be brought into starker relief if we start by comparing two systems: one trained to answer questions about baseball and the other trained to give baking advice. While it is true that the guts of both of these models uses next token prediction, we can sensibly ask why the first system is very good at producing answers about who pitched no hitters in the 1970s but not very good at telling you how to whisk an egg. Appealing to next token prediction alone provides no clear way to answer these straightforward questions. The impressive accomplishment of LLMs is that they produce whole paragraphs of text relevant to their specific contexts. In both the baseball and baking cases, the output is achieved by both next token prediction *and* token network mapping. In what follows, we'll expand on and explain this basic idea and tell you why it's important that we not adopt the hyper-reductive stance of "just next token".

*Just Next Token Prediction*

What do LLMs do? In some sense, they answer questions, fix code, write stories, and so on. But when we are being a bit more hard-nosed, might these be metaphors? What do LLMs *really* do?

In simple terms, an LLM like ChatGPT is a system that is trained in such a way that when it receives a prompt it begins generating text in reply chunk by chunk by performing statistical operations on arrays of numbers that correspond to words and their distributional features so as to predict a further chunk based on the existing chunks. LLMs are very clever things in no small part because of their attention heads and large context windows. LLMs come to learn interesting features about sentences such as how relevant other words are to resolving anaphoric terms or whether a word is being used as a verb or a noun. But, in some sense, this is all in the service of producing a next word that gets mathematically rewarded. From this vantage, it's understandable why one might say that such a system is "just a next word predictor". (In what follows, the difference between words and tokens will not matter, so we will slide over the difference for readability.) Let's call proponents of "just next word" JNPers.

A nearby claim is that LLMs are "stochastic parrots". It's worth flagging that this is a distinct claim from the claim that LLMs are mere next word predictors. The term comes from Bender et al. (2021) and aims to capture the following thought: much like a parrot who says, "Polly wants a cracker," LLMs produce strings that humans can interpret, but (if the objection is correct), the LLM itself has no knowledge of the meaning of the string it produces. It might be that a just-next-word predictor is a stochastic parrot if just-next-wording entails being non-sensitive to meaning. But it isn't obvious that the entailment holds. It's possible that a next-word-predicting mechanism could be a meaning user, at least in principle. And the other direction seems even more plausible still: parrots aren't next word predictors, they are sound-mimickers. But if anything deserves that charge of parrotry, it's parrots. The present paper is not focused on whether LLMs are meaning users in the sense of being sensitive to the semantics of the strings they take as input or give as output (see Titus 2024 for further discussion). It seems to us that much of that debate will boil down to how convinced one is that distributional semantics is a viable theory of *meaning*. The present paper is focused on the claim that what LLMs *do*, when we are being hard-nosed and serious, is simply predict the next word/token.

Such a claim isn't a labeling exercise, a mere semantic dispute, or only of interest to philosophers. We will see below that there are explanatory costs and benefits that come with taking the "next token's eye view" as compared to a higher-level style of description and explanation, but the issue can be motivated in simple terms. Suppose everyone is convinced that LLMs and other forthcoming AI systems are just next token predictors. If correct, it's hard to see why we should think they might pose an existential threat. It's hard to see how we could lay blame for biased outputs at their feet. It's hard to see how they could be viewed as anything other than a very clever gimmick. But if they are more than next word predictors — if they are, for example, advice givers, question answerers, trip planners, and so on, what we expect of them, and the standards to which we hold them, shift in important ways. Again, below we will dig into deeper explanatory differences, but we hope this helps motivate the issue at the outset.

Let us returning to next word prediction and what LLMs do. One way to get started is to see just how far we can push an idea. So, why not adopt a *really* hard-nosed, reductionist view and say LLMs aren't even next word predictors, they are just number crunchers. After all, what's *really* going on (the thought continues) is a lot of linear algebra. Why even go so far as saying that LLMs do anything with words or tokens at all?

It's worth exploring why no one is tempted to say this. The important take-away will be that reductionism always faces a certain kind of pressure to say when one has reduced far enough. Going down to number crunchers is to go too far. But by similar lights, reducing to 'just next token' is, we think, also to go too far. So let's start with an argument for rejecting the view that LLMs are "just number crunchers".

Return to the model's objective: next token prediction. Contemporary LLMs are optimized on next word or next token. When the system is working as designed and all is going well, it predicts a next word. That is, there is a task or objective the system is given – predict a hidden word that comes next in a sequence – and, during training, it is rewarded or punished (in a mathematical sense) depending on whether it correctly or incorrectly predicts the hidden word. To achieve this, the system undergoes a process of weight adjustment and a great deal of number crunching. But in light of its token-level objective and pattern of rewards and punishments, it is correct to say that the LLM is a next token predictor and not just a number cruncher.

Is this where we should stop – more than a number cruncher but no more than a next token predictor?

We don't think so and we will offer two arguments. First, we will offer an argument from change in function. Given the sort of thinking that moves one away from number cruncher and towards next token prediction, there are reasons one should go a little further. Second, we will offer an argument on the basis of explanatory gains and losses that draws an analogy from explanation in biology. Both arguments speak in favor of seeing LLMs as more than just next token predictors.

*Functions and Changing Functions*

The function of an object or structure is, according to one influential approach, what it is designed to do. If a thing's function were set in stone once and for all, perhaps we should treat LLMs as next word predictors and nothing more. But functions can change – things designed or selected to do one thing can be recruited to do another and can come to have a new function. For example, Play-Doh was originally a product introduced as a wallpaper cleaner, but as children began to play with it, it came to be seen as a toy. Now we know Play-Doh as a toy and not wallpaper cleaner. But why? Shouldn't we say that wallpaper cleaner is *being used as* a toy but isn't *really* a toy? No. History might have unfolded in such a way that the manufacturer continued to sell the product as a wallpaper cleaner and continued to refine the product for the purposes of cleaning only, but that's not, in fact, how it went. The very same product began being sold as a toy *and optimized for being a toy*. The product was tweaked and adjusted in light of this new use — adding colors for example and creating fun molds. What once was wallpaper cleaner, through changes in *both* use and development, came to be a toy.

Of course, using something *as* something else doesn't make it so, so we need to proceed cautiously. For example, WD-40 is used by many as a lubricant and rust remover. But WD-40 was designed to displace water from machine parts (hence the "WD").

Presumably, in testing and development, the substance was optimized on how well it could displace water without otherwise damaging the relevant machine parts. We know it went through some development, as we are now on the 40th formula, but there isn't very good evidence that the company has since changed the product in order to make it better at lubricating and removing rust. The company *has* released variations such as WD-40 BIKE, but the base product seems to have stayed largely the same since its introduction. But people *use it* regularly as a lubricant. So is it a lubricant as well as a water displacer? Here, we should say "no" given its design history. It is indeed good at lubricating, but it is "just a water displacer". Unlike Play-Doh, the trajectory of WD-40 has not been guided by its use as a lubricant and rust remover, and this makes all the difference to what it is.

With those two examples in mind, return to LLMs. LLMs, in their unsupervised training phase, are guided by a next word prediction objective. And as noted above, this is achieved through the adjustment of weights and a lot of vector manipulation. By adjusting the weights and crunching numbers with the objective of predicting the next word, we have a next word predictor. With respect to number crunching versus predicting next words, LLMs are more like PayDoh and less like WD-40 and should be seen as more than mere number crunchers.

That helps get us beyond number-cruncher and back up to next-word-predictor, but should we go further? Plausibly, we can say "yes" at this juncture given that LLMs go through fine-tuning and through reinforcement learning with human feedback in order to be optimized to answer questions, provide advice, make jokes, and so on. A model might be trained to produce specialist outputs such as medical reports and might be trained with an eye toward performance improvement such as increased consistency. Turning to reinforcement learning through human feedback (RLHF), a system might be trained with the objective of providing outputs that answer questions to the satisfaction of a user or that of providing outputs that are agreeable in tone. Typically, there is a lot more training beyond the initial unsupervised learning stage and that's enough to make the system something more than just a next word predictor. So, once again, more like Pay-Doh than like WD-40. So the JNPer is not on stable footing. We don't stop at "number cruncher" because of the design history and development. Similarly, we shouldn't stop at "just next word".

*Explanatory gains and losses: a lesson from biology*

We wish to offer a second argument against the JNPer. Here we turn to the potential explanatory gains of going beyond the JNP view. We proceed by analogy with the gene's eye view in biology. The gene's eye view stakes out a particularly reductionist stance in a debate over the units (or levels) of selection. The debate involves how selection operates on populations—whether the process unfolds on the level of genes, organisms, groups, or all of the above. Because life is organized hierarchically, and multicellular life relies on cell-to-cell signaling and genetic transmission, there are influential arguments put forward that claim we can reduce the operation of evolution to the piecewise assembly and careful tuning of genes. The so-called "gene's eye view" of evolution, first articulated by George Williams (1966) and popularized by Richard Dawkins (1976) and Daniel Dennett (see especially 1995), claims that genes and only genes are the relevant units of selection.

As tempting as this reductionist view may be, it faces serious problems that are well-documented in the literature (See Agren 2021 for a detailed overview of these issues). Genes need to be assembled into genomes that are functionally integrated, and these integrated genomes are fragile in all sorts of complex ways that are not obvious, predictable, or explainable by focusing only on the individual genetic ingredients. For instance, cases of selfish genetic elements involve conflicting selection pressures. There is selection at the level of the genome to preserve its functional integrity, yet there is also selection at the gene level for the element to make as many copies as possible. Proliferation of selfish genetic elements can–and often does–disrupt the functionality of the genome's ability to manufacture and regulate proteins. If we want to understand the system, we need to identify selection at both levels, and recognize the higher-order organizational structure at the genome level. Focusing only on competition between genetic elements blinds us to the actual structure of biological evolution.

From a developmental perspective, the gene's eye view faces even stiffer challenges. Complex multicellular organisms have developmental trajectories that involve much more than gene action. Environmental resources and spatial organization play crucial roles in development and it is, at minimum, extremely difficult (but probably impossible) to reduce these factors to gene action. So the gene's eye view, it is generally agreed, no longer provides sufficient explanatory resources in either an evolutionary or a developmental context. Understanding evolutionary patterns of changes in variation across generations requires attention to more than just strands of DNA passed from one generation to the next. In development, DNA is just one of the many resources required to produce an organism.

The relevant point is not simply that some proposed reduction in one area of science fails and so thinking on LLMs should follow suit. The lessons run deeper, and to see why, it's worth taking a moment to diagnose the temptation towards reductionism in both biology and with respect to LLMs.

We suspect the appeal of the reduction can be explained by focusing on assemblage. If only we could explain the *building process*, the thought goes, we'd explain everything that needs explaining. More specifically, there was an enthusiasm for focusing on the mechanisms for transmitting DNA by proponents of the gene's eye view – explain this and you will have explained everything.

But this view hasn't held sway. The debate over the gene's eye view led to some important theoretical advances for thinking about evolutionary dynamics that takes us outside of the grips of reductionism. One early and valuable proposal, put forward by the biologist and philosopher David Hull (2001), involves a distinction between replicators and interactors. *Replicators* are the units that are copied and help explain cross generational transmission of traits and the reproduction of organisms. *Interactors* are the units that engage causally with the world and make possible the differential success of their embedded replicators. Whereas the proponents of the gene's eye view argued for a focus only on replicators, Hull argued that there is an important place for each kind of unit in evolution. It turns out that life evolved to use replicators as an important component of biological inheritance and understanding how replicators encode and transmit information is crucial to understanding the causal workings of the cell. Yet the interactors are also indispensable. Oyama (2000) and others pursue this point further arguing organisms are built by much more than genes and reflect a unity of purpose or functional operation that plays the key role in the living, dying and reproducing that drives evolution by natural selection. In effect, we have a distinction between the tokens that are used to assemble important higher order functional units and the scope of functional organization that identifies the higher order units.

Returning now to LLMs, JNPers look to be engaged in a project that bears an important similarity to the gene's-eye-view approach. While JNPers are correct that next token prediction is the *mechanistic method* that *assembles* the sentences and paragraphs, assembly does not exhaust the relevant structural information that is reflected in the final assembled product. This limits the ability of JNPers to explain how LLMs learn to compose sentences and paragraphs.

Further, there is a deep formal connection between reinforcement learning and evolutionary dynamics that makes this comparison between LLMs and units of selection more than just superficially similar. To see this, consider a simple reinforcement learning model. A model learning by reinforcement selects a response–a strategy in a game, a token to slot in a string next–randomly. If the response is successful, the model proportionally increases the learning weight associated with the response. As learning progresses, best responses accumulate more and more weight and the model tends to select these responses almost all the time.

To make the connection between this simple reinforcement model and evolutionary dynamics, envision the learning weights as proportions of individuals producing the particular responses in a population. The payoffs associated with success would then be a measure of the fitness consequences for an individual's response. Successful individuals produce more of the same type and the proportion of that response increases in the population. As natural selection operates, more and more of the population will be adopting one of the best responses. Instead of changing learning weights, and therefore the chance a model selects a particular response, the evolutionary dynamics change the composition of the population and therefore the chance of randomly encountering an individual producing a particular response.

Just as evolution by natural selection can assemble sets of replicators into functionally organized interactors, reinforcement learning targeted on next token prediction can generate association networks among tokens that form functionally organized sentences and paragraphs. The claim that LLMs are just next token predictors blinds us to the capabilities of reinforcement learning to achieve this benchmark, and undermines our ability to explain how such structural information is learned by the model.

A sign that this higher level structural information–e.g. inter- and intra- sentential syntactic information–is revealed by the expansive attention heads the models use to make next token predictions. Predicting the next token, conditional on the previous token only, is the simplest form of prediction. If LLMs did only this, they would not be very successful at answering questions or telling stories and would perhaps indeed be mere next token predictors. But attention heads expand the scope of prediction by making available information concerning many previous tokens and their interrelations to predict a next token. While the algorithm assembles sentences token by token, the model has learned complex association networks among tokens. Upon first use of a chatbot like ChatGPT, for example, it can feel like a bit of magic that something that is laying down a new token, one token at a time, could possibly produce not only whole sentences that are grammatical, but paragraphs, whole essays, and beyond, all that make sense. It is the association among tokens that allows the LLMs to make such good next token predictions and therefore assemble *sentences* that are *sensible answers to questions*. This association network encodes the higher level structural information that reflects the functional organization between sets of tokens that make up correct sentences and beyond. This is among the information the LLM has learned. Just as evolution assembles complex gene networks into functional genomes, the LLM has effectively learned syntactic and (possibly) semantic structure that can be reproduced by conditioning next token prediction on a bigger and bigger set of previous tokens. This is the sort of information Anthropic (Templeton, et al 2024) is identifying in the LLMs when they are "mapping the mind" and revealing concept clusters and other abstract features encoded in the model. So, in effect, LLMs are not *just* next token predictors, but next token predictors *and* token network mapping devices. By leveraging the rich networks of connections between tokens, the LLMs can sequentially assemble coherent sentences and paragraphs.